\documentclass[sigconf, nonacm, 9pt]{acmart}

\usepackage{microtype}
\usepackage{graphicx}
\usepackage{booktabs}
\usepackage{hyperref}
\usepackage{array}
\usepackage{multirow}
\usepackage{algorithm}
\usepackage{algorithmic}
\usepackage{subcaption}
\usepackage{amsfonts}
\newcommand{\todo}[1]{#1}
\newcommand{\edd}[1]{#1}

\AtBeginDocument{%
  \providecommand\BibTeX{{%
    \normalfont B\kern-0.5em{\scshape i\kern-0.25em b}\kern-0.8em\TeX}}}

\begin{document}

% \title[SWIM: Selective Write-Verify for Computing-in-Memory Neural Accelerators]{SWIM: Selective Write-Verify for Computing-in-Memory \\Neural Accelerators\\ \small{\ } \\ \small{\ } \\  \small{\ } \\  }
\title[SWIM: Selective Write-Verify for Computing-in-Memory Neural Accelerators]{SWIM: Selective Write-Verify for Computing-in-Memory \\Neural Accelerators }

\author{Zheyu Yan} 
\email{zyan2@nd.edu}
\affiliation{%
  \institution{University of Notre Dame}
  \country{}
}
\author{Xiaobo Sharon Hu}
\email{shu@nd.edu}
\affiliation{%
  \institution{University of Notre Dame}
  \country{}
}
\author{Yiyu Shi}
\email{yshi4@nd.edu}
\affiliation{%
  \institution{University of Notre Dame}
  \country{}
}


\begin{abstract}
  Computing-in-Memory architectures based on non-volatile emerging memories have demonstrated great potential for deep neural network (DNN) acceleration thanks to their high energy efficiency. However, these emerging devices can suffer from significant variations during the mapping process (\emph{i.e.}, programming weights to the devices), and if left undealt with, can cause significant accuracy degradation. The non-ideality of weight mapping can be compensated by iterative programming with a write-verify scheme, \emph{i.e.}, reading the conductance and rewriting if necessary. In all existing works, such a practice is applied to every single weight of a DNN as it is being mapped, which requires extensive programming time. In this work, we show that it is only necessary to select a small portion of the weights for write-verify to maintain the DNN accuracy, thus achieving significant speedup. We further introduce a second derivative based technique {\em SWIM}, which only requires a single pass of forward and backpropagation, to efficiently select the weights that need write-verify. Experimental results on various DNN architectures for different datasets show that {\em SWIM} can achieve up to 10x programming speedup compared with conventional full-blown write-verify while attaining a comparable accuracy. 
\end{abstract}

\maketitle

\section{Introductions}

Deep Neural Networks (DNNs) have surpassed human performance in various perception tasks including image classification, object detection, and speech recognition. Deploying DNNs on edge devices such as automobiles, smartphones, and smart sensors is a great opportunity to further unleash their power. However, because edge platforms have constrained computation resources and limited power budget, employing CPUs or GPUs to implement computation-intensive DNNs on them is a great challenge.

%Conventional von-Neumann edge DNN accelerators generally use on-chip process elements (PEs) for computation and turn to off-chip storage for data (\emph{e.g.}, weight) preservation. Busy data transfer in different levels of memory across static random-access memory (SRAM) in PEs and off-chip storage induces great energy and latency overhead thus hinders accelerator performances. This phenomenon called \emph{the memory wall} is an universal issue of the conventional von-Neumann architecture.

Non-volatile Computing-in-Memory (nvCiM) DNN accelerators~\cite{shafiee2016isaac} offer a great opportunity to edge applications by reducing data movement with an in-situ weight data access scheme~\cite{sze2017efficient}. By making use of emerging non-volatile memory (NVM) devices \edd{(\emph{e.g.}, resistive random-access memories (RRAMs), ferroelectric field-effect transistors (FeFETs) and phase-change memories (PCMs)),} nvCiM can achieve higher energy efficiency and memory density compared with conventional MOSFET-based designs~\cite{chen2016eyeriss}.
However, NVM devices suffer from various non-idealities, especially device-to-device variations due to fabrication defects and cycle-to-cycle variations due to the stochastic behavior of devices. If not properly handled, the weights actually mapped to the devices could deviate significantly from the expected values, leading to large performance degradation.

Different strategies have been proposed to tackle these issues. Noise-aware training~\cite{jiang2020device} and uncertainty-aware neural architecture search~\cite{yan2020single, yan2021uncertainty, yan2021radars} aim at fortifying DNNs so that their performance remains mostly unaffected even in the presence of device variations. However, these methods are not economical because they require re-training DNNs from scratch and cannot make use of existing pre-trained models. On-chip in-situ training~\cite{yao2020fully}, on the other hand, directly fine-tunes the DNNs through additional training after they are mapped to nvCiM platforms so that the impact caused by weight variations during mapping can be alleviated. This method is quite effective but requires extra hardware to support backpropagation and weight update. In addition, it requires iterative training which involves multiple cycles of write for each weight being updated and can take quite some time.  

As such, a widely adopted practice today is write-verify, which applies iterative write and read (verify) pulses to make sure that the weights eventually programmed into the devices differ from the desired values by an acceptable margin. Write-verify can reduce the weight deviation from the ideal value to less than 3\% and the DNN accuracy degradation to less than 0.5\%~\cite{shim2020two}. However, write-verify is time-consuming because each weight value needs to be written-verified individually. Programming even a ResNet-18 for CIFAR-10 to an nvCiM platform can take more than one week~\cite{shim2020two}. Considering that the programming time grows linearly \emph{w.r.t.} the number of parameters in the DNN model and many state-of-the-art models have far more weights than ResNet-18, an interesting question is, \textbf{whether we really need to write-verify every weight of a DNN when mapping it to an nvCiM platform.}

In this work, we show that the answer to \todo{the} question is \textbf{NO}. It is in fact only necessary to write-verify a small portion of the weights to attain an accuracy very close to that assuming ideal mapping, and as such, the programming time for nvCiM platforms can be drastically reduced. Specifically, we propose \underline{S}elective \underline{W}rite-verify for computing-\underline{I}n-\underline{M}emory neural accelerators ({\em SWIM}). Different from the vanilla write-verify scheme that performs write-verify for all the weights to be mapped, inspired by~\cite{lecun1990optimal}, {\em SWIM} uses second derivatives of the weights as an indicator to select only a small portion of the \emph{sensitive} weights to write-verify. In addition, considering that straight-forward computation of second derivatives through finite difference method is extremely expensive, we devise a forward and backpropagation scheme similar to what is in gradient computation, which only takes a single pass, to get all the \edd{second derivative data}. Experimental results on MNIST, CIFAR-10, and Tiny ImageNet show that {\em SWIM} can achieve up to 10x, 5x, and 9x programming speedup compared with the conventional approach of writing-verifying all the weights, a magnitude based selective write-verify heuristic, and a state-of-the-art in-situ training, respectively. 
To the best of our knowledge, this is the first work that establishes the concept and verifies the effectiveness of selective write-verify framework for programming nvCiM neural accelerators.
\section{Related Works}

\subsection{Crossbar-based Computing Engine}\label{sec:2.1}
% \begin{figure}[ht]
% %\vskip 0.2in
% \begin{center}
% \centerline{\includegraphics[trim=0 150 550 0, clip, width=0.4\linewidth] {figures/crossbar.pdf}}
% \caption{Illustration on crossbar architecture. The input is fed horizontally and multiplied by weights stored in the NVM devices at each cross point. The multiplication results are summed up vertically and the sum serves as an output.}
% \vspace{-0.5cm}
% % \todo{This figure is generated from testcode/CIM/9874/plot/crossbar.pptx}
% \label{fig:crossbar}
% \end{center}
% \end{figure}

Crossbar array is a key component of nvCiM DNN accelerators. A crossbar array can be considered as a processing element for matrix-vector multiplication where matrix values (\emph{e.g.}, DNN weights) are stored at the cross point of each vertical and horizontal line with resistive emerging devices such as RRAMs, FeFETs, and PCMs, and each vector value (\emph{e.g.}, DNN inputs) is propagated through horizontal data lines. The calculation in crossbar is performed in the analog domain but additional peripheral digital circuits are needed for other key DNN operations (\emph{e.g.}, pooling and non-linear activation), so digital-to-analog and analog-to-digital converters are used between different components.

\edd{Resistive crossbar arrays suffer from various sources of variations and noises. Two major ones include spatial variations and temporal variations. Spatial variations result from fabrication defects and have both local and global correlations. NVM devices also suffer from temporal variations due to the stochasticity in the device material, which causes fluctuations in conductance when programmed at different times. Temporal variations are typically independent from device to device and are irrelevant to the value to be programmed~\cite{feinberg2018making}.  
% This contributes to the majority of the impact of variations. 
In this work, as a proof of concept, we focus on the impact of temporal variations in the programming process on DNN performance. Temporal variation makes the programmed resistance of a device to deviate from what is expected. The proposed framework can also be extended to other sources of variations with modification.}  

\subsection{Handling Variations in Weight Mapping}
Various approaches have been proposed to deal with the issue of device variations on nvCiM DNN accelerators. Here we briefly review the two most common ones that do not require training a new model from scratch. 

On-chip in-situ training, which fine-tunes a trained DNN directly on nvCiM platforms, can recover model performance in a few iterations if device variations are small. In each iteration, the forward and backpropagation process is performed on-chip under the impact of device variations, and each weight is updated by applying voltage pulses to the corresponding device. The number of write pulses is determined by the gradient of that weight. Such a scheme requires extra hardware support for training and can be quite time-consuming due to multiple iterations of write for each weight. \edd{More recent works propose to only fine-tune the fully connected layers of DNN models~\cite{yao2020fully}, but the effectiveness of this method on large models is unclear.}

In write-verify, an NVM device is first programmed to an initial state using a pre-defined pulse pattern, then the value of the device is read out to verify if its conductance falls within a certain margin from the desired value (\emph{i.e.}, if its value is precise). If not, an additional update pulse is applied, aiming to bring the device conductance closer to the desired one. This process is repeated until the difference between the value programmed into the device and the desired value is acceptable. The process typically requires a few iterations. More seriously, write-verify is performed individually for each weight as it is being mapped to a device. Therefore, writing-verifying a large number of NVM devices requires much longer programming time than writing-without-verify which is done in parallel.

%Noise-aware training of DNNs is an effective approach to reduce the impact of device variations. These methods view the impact of device variation as noise.
%Noise-injection-based training~\cite{jiang2020device} samples an instance of noise from the device variation model and injects it into the DNN model in each iteration of training so that the model can learn to operate in noisy environments.
%Bayesian Neural Network (BNN)-based training~\cite{gao2021bayesian} integrates the noise model into the variational model of BNNs so a robust DNN model can be obtained by direct training of the BNN. Noise-aware training can offer DNN models that are statistically robust against the impact of device variations even if the device variation is significant. However, they require more training iterations (usually 10x) to converge than training without noise and starting from a pre-trained models is not helpful~\cite{jiang2020device}.

\section{\emph{SWIM} Framework}

\subsection{Overview of \emph{SWIM}}

\begin{figure}[t]
%\vskip 0.2in
\begin{minipage}[b]{0.45\linewidth}
    \includegraphics[trim=5 0 30 30, clip, width=\linewidth]{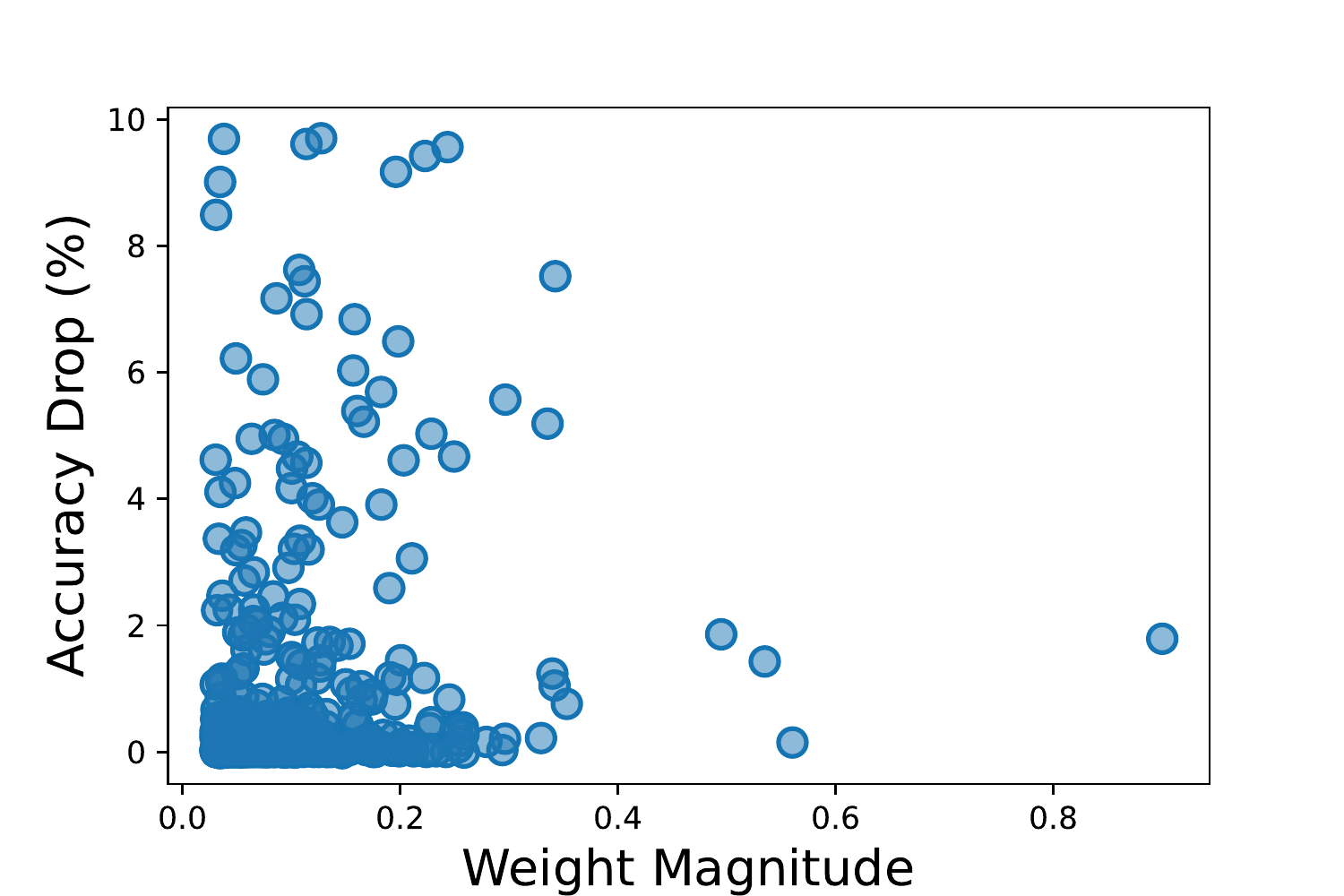}
    \vspace{-0.5cm}
    \subcaption{}
    \vspace{-0.5cm}
    \label{fig:mag}
\end{minipage}
\hspace{0.05\linewidth}
\begin{minipage}[b]{0.45\linewidth}
    \includegraphics[trim=5 0 30 30, clip, width=\linewidth]{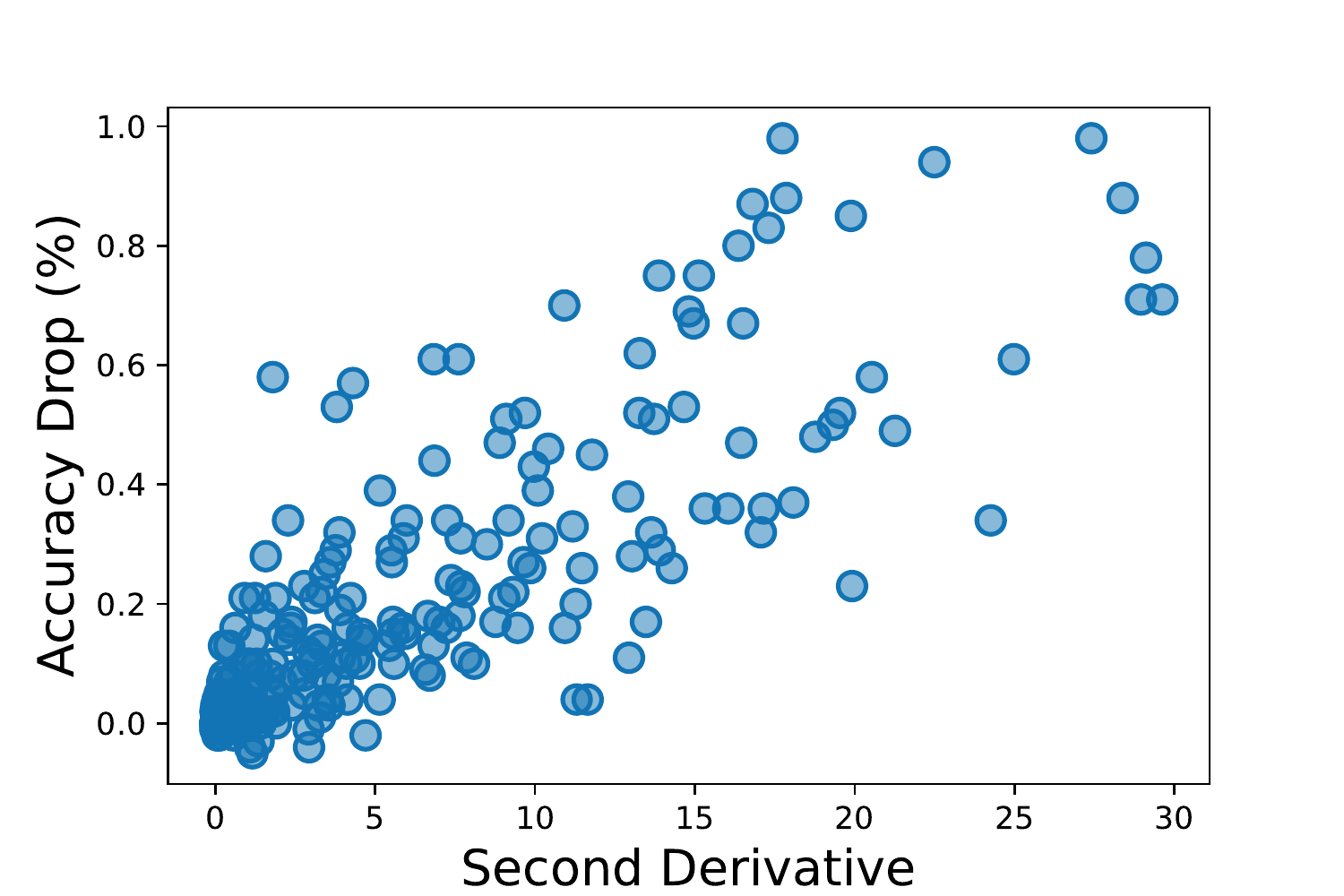}
    \vspace{-0.5cm}
    \subcaption{}
    \vspace{-0.5cm}
    \label{fig:second}
\end{minipage}

\caption{Impact of LeNet weight variation on MNIST: (a) Accuracy drop v.s. weight magnitude, where little correlation exists. (b) Accuracy drop v.s. second derivative of each weight, where strong correlation can be observed.}
\vspace{-0.3cm}
% \todo{This figure is generated from testcode/CIM/9874/plot/crossbar.pptx}
\label{fig:correlation}
\end{figure}

In this paper, contrary to the practice of all existing works that perform write-verify for every weight of a DNN, we establish and explore answers to the following problem. 

\noindent {\bf Selective Write-Verify}: Given a DNN architecture with weights $\mathbf{W_0}$ and a maximum acceptable accuracy drop $\delta A$, identify the smallest subset $\mathbf{W_s}\subseteq \mathbf{W_0}$ so that, when mapping the DNN to nvCiM platforms, by only writing-verifying weights in $\mathbf{W_s}$, the deployed network can have an accuracy no less than $\delta A$ below that of the original network. 

\todo{One important feature for NVM devices is that the read process takes much shorter time than write \cite{shafiee2016isaac}, especially for RRAMs and FeFETs. As such, reading the values of weights programmed into the devices and evaluating the corresponding accuracy of the DNN takes negligible amount of time compared with the write-verify process.} We can leverage this feature to develop a heuristic approach to address the selective write-verify problem through iterative mapping, as shown in Alg.~\ref{alg:swim}. For each weight in $\mathbf{W_0}$, we can first evaluate the impact of its variation on the accuracy of the DNN, which is referred to as its {\em sensitivity} in the remainder of this paper. Then we sort all the weights in descending order of {\em sensitivity}, and iteratively write-verify a group $p$ of the weights at a time (called programming granularity in Alg.~\ref{alg:swim}) until the accuracy drop is below $\delta A$. In our experiments, we find that setting $p$ to be $5\%$ of the total number of weights is sufficient to provide the granularity for improving accuracy, while also avoiding too frequent evaluation of the accuracy of the mapped DNN. A critical question now is how to evaluate the {\em sensitivity} of a weight, which will be discussed in the next section. 

\begin{algorithm}[h]
\caption{SWIM~($\mathbf{W_0}$ , $\mathcal{Z}$,  $A$, $\delta A$, $\mathbf{D}$, $p$)}
\begin{algorithmic}[1]\label{alg:swim}
\STATE // INPUT: A trained DNN architecture $\mathcal{Z}$ with weights $\mathbf{W_0}$, original DNN accuracy $A$, the maximum accuracy drop allowed $\delta A$ after mapping to nvCiM, training dataset $\mathbf{D}$, and programming granularity $p$;

\STATE Program all weights in $\mathbf{W_0}$ based on their locations in $\mathcal{Z}$ to the nvCiM platform;
\STATE Calculate the \emph{sensitivity} of all the weights;
\STATE Sort $\mathbf{W_0}$ in the descending order of \emph{sensitivity}. 
\FOR{($i=0$; $i < (|\mathbf{W_0}|/p)$; $i++$)}
    \STATE Write-verify the weights $\mathbf{W_0}[i\times p+1: (i+1)\times p]$ based on their positions in $\mathcal{Z}$;
    \STATE Evaluate the accuracy $\tilde A$ of the mapped network on $\mathbf{D}$;
    \IF{$A-\tilde A \leq \delta A$}
        \STATE Break;
    \ENDIF
\ENDFOR
\end{algorithmic}
\end{algorithm}

\subsection{Sensitivity Analysis}\label{sec:sensitivity}

Our goal is to find a way to evaluate the {\em sensitivity} for each weight. 
Intuitively, one would think that the magnitude of a weight would be a good indicator for 
{\em sensitivity}, and that, the larger the weight is, the more impact on the accuracy it would have 
when it is being perturbed. Unfortunately, our preliminary studies show that this is not the case.
% As discussed in Section~\ref{sec:2.1} and Section~\ref{sect:model}, 
From experimental results in~\cite{feinberg2018making}, we assume a model where the amount of variances of NVM devices are independent of the 
value to be programmed. We perturb each weight in LeNet with the same additive Gaussian noise based on \cite{yao2020fully} 
and evaluate the corresponding drop in the DNN accuracy for perturbing each weight, averaged over 100 Monte Carlo runs. 
From Fig.\ref{fig:mag}, 
% \edd{where each point represents the experiment of perturbing one weight,}
we can see that there is very weak correlation, 
if any, between the magnitude of weights and the accuracy drop that their variations cause. 
This observation is further confirmed in the experimental section, where 
we show that magnitude based selection approach would not yield good results. Below, 
we present a rigorous mathematical analysis, to establish a quite effective metric that can reflect the {\em sensitivity} of a weight. 

As existing DNN optimization engines all map accuracy maximization to the 
minimization of a loss function, there is a strong correlation between the impact 
of a weight's variation on accuracy and that on the loss function. As such, we 
resort to evaluating the {\em sensitivity} based on the latter.  

For a DNN with a given labeled training dataset, loss $f$ is a function of 
a vector $\mathbf w$ formed by all the weights. Assume that the training is completed 
and the optimal weights identified are $\mathbf{\tilde{w}}$. With small variations of 
the weights around $\mathbf{\tilde{w}}$, \emph{i.e.}, $\mathbf{w} = \mathbf{\tilde{w}}+\mathbf{\Delta w}$, 
one can perform Taylor expansion on $f$ as follows: 
\begin{equation}
    f(\mathbf{w}) = f(\mathbf{\tilde{w}}) + \frac{\partial f}{\partial \mathbf{\tilde{w}}} \Delta\mathbf{w} + \frac{1}{2} \Delta\mathbf{w}^T \mathcal{H}(\mathbf{\tilde{w}}) \Delta\mathbf{w} + o(\Delta \mathbf{w}^{3}) \label{eq:1}
    \vspace{-0.2cm}
\end{equation}
where we use the compact notation $\frac{\partial f}{\partial \mathbf{\tilde{w}}}$ to 
represent $\frac{\partial f}{\partial \mathbf{w}}|_{\mathbf{w}=\mathbf{\tilde{w}}}$. Similar notation 
will be used throughout the paper. 
$\mathcal{H}(\mathbf{w})$ is the Hessian of $\mathbf{w}$ defined as 
\vspace{-0.1cm}
\begin{equation}
    \mathcal{H}(\mathbf{w}) = \left[\begin{array}{ccc}
\dfrac{\partial^2 f}{\partial w_1^2} & \cdots & \dfrac{\partial^2 f}{\partial w_1 \partial w_n} \\
\vdots & \ddots & \vdots \\
\dfrac{\partial^2 f}{\partial w_n \partial w_1} & \cdots & \dfrac{\partial^2 f}{\partial w_{n}^2}
\end{array}\right]
% \vspace{-0.2cm}
\end{equation}
with $n$ being the total number of weights, \emph{i.e.}, length of $\mathbf{w}$.

As the neural network is trained to convergence through gradient descent, we have 
$\frac{\partial f}{\partial \mathbf{\tilde{w}}}=\mathbf{0}$. Accordingly, based on Eq.~\ref{eq:1} 
the change in the loss function $\Delta f(\mathbf{w})$ brought by the weight variation $\Delta \mathbf{w}$ around $\mathbf{\tilde{w}}$
can be expressed as 
\begin{equation}
    \Delta f(\mathbf{w}) \approx  \frac{1}{2} \Delta\mathbf{w}^T \mathcal{H}(\mathbf{\tilde{w}}) \Delta\mathbf{w}  \label{eq:2}
    \vspace{-0.1cm}
\end{equation}
where we have ignored the higher-order terms.  

Recall that $\Delta \mathbf{w}$ is device-specific and independent of the magnitude of $\mathbf{w}$~\cite{feinberg2018making}.
% (detailed discussion in Section~\ref{sec:2.1}). 
It is now clear that for a trained model, it is in fact the Hessian that plays 
a critical role in {\em sensitivity}. Unfortunately, the number of elements in Hessian 
is quadratically proportional to the number of weights. For example, a small neural network 
with one million weights ($10^6$) would require a Hessian with one trillion ($10^{12}$) elements, which is computationally impractical to evaluate. 

To explore potential simplification, we notice that Eq.~\ref{eq:2} can be expressed as 
% \begin{equation}
%     \mathcal{E} \approx l(\mathbf{w_T}) + \sum^{|\mathbf{w}|}_{i=1}\frac{\partial f}{\partial w_i} \mathcal{N}_i + \sum^{|\mathbf{w}|}_{i=1}\sum^{|\mathbf{w}|}_{j=1}\frac{\partial^2 f}{\partial w_i \partial w_j} \mathcal{N}_i\mathcal{N}_j\\
% \end{equation}
\vspace{-0.2cm}
\begin{align}
\begin{split}
    \Delta f(\mathbf{w}) & \approx \frac{1}{2} \sum^{n}_{i=1}\sum^{n}_{j=1} \mathcal{H}_{ij} \Delta w_i \Delta w_j\\
    & =  \frac{1}{2} \sum^{n}_{i=1} \mathcal{H}_{ii} (\Delta w_i)^2 + \frac{1}{2} \sum^{n}_{i\neq j} \mathcal{H}_{ij} \Delta w_i \Delta w_j\label{eq:taylor}
\end{split}
\vspace{-0.2cm}
\end{align}
where $\Delta w_i$ is the $i^{th}$ element of $\Delta \mathbf{w}$ and $\mathcal{H}_{ij}$ is the element in the $i^{th}$ row and $j^{th}$ column of $\mathcal{H(\mathbf{\tilde{w}})}$. To simplify Eq.~\ref{eq:taylor}, we assume that the change in the loss function caused by the variations in multiple weights is approximately the sum of those caused by each weight. As such, we only have to deal with 
one weight variation at a time. In this case, the cross terms can be discarded since either $\Delta w_i$ or $\Delta w_j$ is zero when $i\neq j$, and we have   
\begin{equation}\label{eq:final}
   \Delta f(\mathbf{w}) \approx \frac{1}{2} \sum^{n}_{i=1} \mathcal{H}_{ii} (\Delta w_i)^2 =  \frac{1}{2} \sum^{n}_{i=1}\frac{\partial^2 f}{\partial \tilde{w}_i^2} \Delta w_i^2 
   \vspace{-0.2cm}
\end{equation}
where $\tilde{w}_i$ is the $i^{th}$ element of $\mathbf{\tilde{w}}$. Extensive experimental study is conducted to confirm this approximation is acceptable.

Eq.~\ref{eq:final} suggests that we only need to obtain the second derivative of each weight $\frac{\partial^2 f}{\partial \tilde{w}_i^2}$ to evaluate the impact of weight variation on loss. 
By writing-verifying a weight $\tilde{w}_i$, 
we are essentially reducing $\Delta w_{i}$. Therefore, it is 
apparent that we shall assign higher priority to reduce the variation of those weights with higher second derivatives $\frac{\partial^2 f}{\partial \tilde{w}_i^2}$. In other words, the second derivative can be used 
as a good {\em sensitivity} metric for {\em SWIM}. The effectiveness of this metric is confirmed in Fig.~\ref{fig:second}, where now with the same setting as in Fig.~\ref{fig:mag} strong correlation can be observed between the accuracy drop after a weight is perturbed and the second derivative of that weight (Pearson Correlation Coefficient being 0.83). 

Finally, when two weights have the same second derivative, we use their magnitudes as the tie-breaker: the larger one will have a higher priority.

% From the noise model, the noise on each weight is sampled from the same distribution, so the term can be simplified to:
% \begin{equation}
%     \mathcal{R} \approx \mathcal{N}^2 \sum^{M}_{i=1}\frac{\partial^2 f}{\partial w_i^2} \\
% \end{equation}

% As $\mathcal{N}^2$ is always positive, to improve the robustness, we want to minimize:
% \begin{equation}
%     \sum^{M}_{i=1}\frac{\partial^2 f}{\partial w_i^2}
% \end{equation}

%  SWIM takes in a neural network model $m$, a training dataset $\mathbf{D}$ and a sparsity target $p$

\subsection{Second Derivative Calculation}
One straightforward way to compute second derivative is to use finite difference method, \emph{i.e.}, 
\begin{equation}
   \frac{\partial^2 f}{\partial \tilde{w}_i^2}  \approx \frac{f(\tilde{w}_i+\Delta w)-2f(\tilde{w}_i)+f(\tilde{w}_i-\Delta w)}{(\Delta w)^2}
   \vspace{-0.2cm}
\end{equation}
where $\Delta w$ is a small positive number. However, in order to get $f(\tilde{w}_i+\Delta w)$ and $f(\tilde{w}_i-\Delta w)$, two passes of forward propagation are needed, after replacing $\tilde{w}_i$ with $\tilde{w}_i+\Delta w$ and $\tilde{w}_i-\Delta w$, respectively. For a network with one million weights, this requires two million passes of forward propagation. 

Inspired by how the gradients of all the weights are efficiently computed through a single pass of forward and backpropagation based on the chain rule and the chain rule approximation of second derivatives presented in~\cite{lecun1990optimal}, below we present a method that can obtain second derivatives of all the weights in a similar way. 

% We use l2 loss as an example to illustrate the idea, and the same process holds for other loss terms including cross entropy~\cite{lecun1990optimal}. 
Let us start with the last fully connected (FC) layer of a DNN. The computation there can be expressed as
\begin{equation}
 \begin{split}
     \mathbf{P} = g_a(\mathbf{I}),\  \mathbf{O} = \mathbf{W}\cdot \mathbf{P}
 \end{split}
 \vspace{-0.3cm}
 \end{equation}
 where $g_a$ is the activation function of the previous layer. $\mathbf{I}$ is the input vector to the activation function. $\mathbf{W}$ is the matrix containing the weights between the two layers. $\mathbf{P}$ is the output of the previous layer. $\mathbf{O}$ is the output of the last layer. We did not include the activation of the last layer as it can be merged into the loss function. %This is also equivalent to:
% \begin{equation}
% \begin{split}
%     P_j = g_a(I_j)\\
%    O_j = \sum_{i=1}^{|\mathbf{P}|}{W_{ji}P_i}
% \end{split}
% \end{equation}
% where $I_i$ is the $i^{th}$ element of the input vector $\mathbf{I}$ and $g_a(\cdot)$ is a non-linear activation function. $P_i$ is the $i^{th}$ element of intermediate result $\mathbf{P}$.  $W_{ji}$ is the element in $j^{th}$ row and $i^{th}$ column of the weight matrix $\mathbf{W}$. Note that we did not explicitly write bias of the FC layer out because it is irrelevant in second derivative calculation.

Consider a loss function $f(\mathbf{O})$ and we want to compute the second derivative for weights $\frac{\partial^2 f}{\partial W_{ji}^2}$ and for inputs $\frac{\partial^2 f}{\partial I_i^2}$. The former will be used as {\em sensitivity} and the latter will be used for further backpropagation to previous layers. Since $\mathbf O$ is a function of $\mathbf{W}$ 
and $\mathbf{P}$, we can apply the chain rule of the second derivative as
%\begin{footnotesize}
%\begin{align}
%\begin{split}
%    l & = \sum_{j=1}^{|\mathbf{O}|}{(GT_{j} - O_j)^2}\\
%      & = \sum_{j=1}^{|\mathbf{O}|}{\left(GT_{j} - \sum_{i=1}^{|\mathbf{P}|}{{W_{ji}P_i}}\right)^2}\\
%      & = \sum_{j=1}^{|\mathbf{O}|}{\left(GT_{j}^2 - 2 GT_{j}\sum_{i=1}^{|\mathbf{P}|}{{W_{ji}P_i}} + \sum_{i=1}^{|\mathbf{P}|}{{W_{ji}^2P_i^2}} + %2\sum_{m\neq k}{w_{jm} w_{jk} P_m P_k}\right)}
%\end{split}
%\end{align}
%\end{footnotesize}
%where $O_j$ the $j^{th}$ element of the output $\mathbf{O}$ and 
%$GT_j$ is the $j^{th}$ element of the ground truth value.
% \begin{align}\label{eq:b_w}
% \begin{split}
%     \frac{\partial^2 f}{\partial W_{ji}^2} &= \left(\frac{\partial^2 f}{\partial O_{j}^2}\right)\left(\frac{\partial O_{j}}{\partial W_{ji}}\right)^2+ \left(\frac{\partial f}{\partial O_{j}}\right)\left(\frac{\partial^2 O_{j}}{\partial W_{ji}^2}\right)\\
%   &= \frac{\partial^2 f}{\partial O_j^2}\times P_i^2 
% \end{split}
% \end{align}
\begin{align}\label{eq:b_w}
\begin{split}
    \frac{\partial^2 f}{\partial W_{ji}^2} = \frac{\partial^2 f}{\partial O_{j}^2}\left(\frac{\partial O_{j}}{\partial W_{ji}}\right)^2+ \frac{\partial f}{\partial O_{j}}\frac{\partial^2 O_{j}}{\partial W_{ji}^2} = \frac{\partial^2 f}{\partial O_j^2}\times P_i^2 
\end{split}
\vspace{-0.3cm}
\end{align}
where the second equality comes from the fact that $O_j$ is a linear function of $W_{ji}$ so $\frac{\partial^2 O_{j}}{\partial W_{ji}^2}=0$.
% In backpropagation, the second derivative of its output $\frac{\partial^2 f}{\partial \mathbf{O}^2}$ is calculated by its outgoing layer. When offered second derivative of its output, by the chain rule, the second derivative of the weights in this can be formally written as:
% \begin{equation}\label{eq:b_w}
%     \frac{\partial^2 f}{\partial \mathbf{W}^2} = \frac{\partial^2 f}{\partial \mathbf{O}^2} (\mathbf{P^2})^T
% \end{equation}
% where $T$ is the transpose of a vector and $\mathbf{P^2}$ is the element wise square of $\mathbf{P}$.
Similarly, we can get the second derivative of the input 
\begin{equation}
    \frac{\partial^2 f}{\partial I_i^2} = g_a'(P_i)^2 \sum_{j=1}^{|\mathbf{O}|} W_{ji}^2\frac{\partial^2 f}{\partial O_j^2} - g_a''(P_i) \frac{\partial f}{\partial I_i}
    \vspace{-0.1cm}
\end{equation}

Assume we use ReLU as activation function. Then, $g_a'(P_i) = sign(P_i) = sign(I_i)$ and $g_a'' = 0$. Thus, second derivatives of the input can be expressed as:
\begin{equation}\label{eq:b_a}
    \frac{\partial^2 f}{\partial I_i^2} = sign(I_i) \sum_{j=1}^{\mathbf{|O|}}W_{ji}^2 \frac{\partial^2 f}{\partial O_j^2}
    \vspace{-0.1cm}
\end{equation}
% \begin{equation}\label{eq:b_a}
%     \frac{\partial^2 f}{\partial \mathbf{I}^2} = sign(\mathbf{I}) (\mathbf{W^2})^T \frac{\partial^2 f}{\partial \mathbf{O}^2}
% \end{equation}
% where here multiply is matrix multiplication and $\mathbf{I^2}$ and $\mathbf{W^2}$ are the element-wise squares.
\edd{The backpropagation process of max pooling layers cancels derivatives of the deactivated inputs (\emph{i.e.}, the second derivatives of the deactivated inputs is zero).} For ResNet and other models with skip connections, similar to backpropagation process used to calculate gradients, the second derivatives of different branches are summed up. Convolution layers, average pooling, and batch normalization layers can be cast in the same form as FC layers, so their backpropagation can share the same scheme as that for FC layers.

In summary, to get these second gradients, we simply need to compute the second derivative of the loss functions with respect to the output of the DNN, \emph{i.e.}, $\frac{\partial^2 f}{\partial O_j^2}$.
%\begin{equation}
%     = \left(1 - \frac{O_j}{\sum_j \exp(O_j)}\right) \left(\frac{O_j}{\sum_j \exp(O_j)}\right)
%\end{equation}
For L2 loss, $\frac{\partial^2 f}{\partial O_j^2}=2$. For cross-entropy loss with softmax, 
\begin{equation}
\frac{\partial^2 f}{\partial O_j^2}     = \left(1 - \frac{O_j}{\sum_j \exp(O_j)}\right) \left(\frac{O_j}{\sum_j \exp(O_j)}\right)
\vspace{-0.1cm}
\end{equation} 
We can then follow Eq.~\ref{eq:b_w} and Eq.~\ref{eq:b_a} to backpropagate layer by layer.

% If the loss function is L2 as shown below, we can have 
% \begin{equation}
%     l  = \sum_{j=1}^{|\mathbf{O}|}{(GT{j} - O_j)^2}
% \end{equation}
% where GT is the ground truth value of the input data.
% \begin{equation}
%     \frac{\partial^2 f}{\partial O_j^2} = 2
% \end{equation}
%If the loss function is cross-entropy with softmax as shown below\footnote{Implementations of cross entropy loss may differ and here we show the implementation form PyTorch.}:
%\begin{equation}
%    l(\mathbf{O}, label) = -O_{label}+\log\left( \sum_j \exp(O_j)\right) 
%\end{equation}
%where $label$ is the ground truth class of the input data.
%Its second derivatives is:
%\begin{equation}
%    \frac{\partial^2 f}{\partial O_j^2} = \left(1 - \frac{O_j}{\sum_j \exp(O_j)}\right) \left(\frac{O_j}{\sum_j \exp(O_j)}\right)
%\end{equation}

Note that the first order gradient can be computed as 
\begin{align}\label{eq:g_w}
    \frac{\partial f}{\partial W_{ji}} &= \frac{\partial f}{\partial O_j} \times P_i \\
    \frac{\partial f}{\partial I_i} &= sign(I_i) \sum_{j=1}^{\mathbf{|O|}}W_{ji} \frac{\partial f}{\partial O_j}\label{eq:g_a}
    \vspace{-0.2cm}
\end{align}

Comparing Eq.~\ref{eq:g_w} and Eq.~\ref{eq:g_a} with Eq.~\ref{eq:b_w} and Eq.~\ref{eq:b_a}, we can find that the second derivative only requires an extra multiplication, and the time needed is negligible compared with convolution operations in forward propagation. If implemented efficiently, the second derivative calculation process of {\em SWIM} takes approximately the same amount of time and memory as conventional gradient computation. In addition, unlike gradient computation that needs to be repeated in each iteration of 
gradient descent, in {\em SWIM} only second derivative computation is done only once. 

\begin{table*}[t]
    \centering
    \vspace{-0.3cm}
    \caption{Comparison of accuracy (\%) and normalized write cycles (NWC) between {\em SWIM} and the baselines on LeNet for MNIST under different $\sigma$, the standard deviation specified in Eq.~\ref{eq:noise} before write-verify. Data are collected over 3,000 Monte Carlo runs and reported in mean$\pm$std format. Write cycles are normalized to those needed to write-verify all the weights. NWC $=0.0$ means no write-verify or in-situ training. NWC $=1.0$ for the three write-verify methods corresponds to the conventional method of writing-verifying all the weights. 
    % For different device variations ($\sigma = 0.1$, $0.2$ and $0.3$), {\em SWIM} can always offer much better accuracy under the same number of write-verify cycles compared with the three baselines.
    }
    \vspace{-0.3cm}
    \begin{tabular}{clccccccc}
        \toprule
        \multirow{2}{*}{$\sigma$}     & \multirow{2}{*}{Method} &\multicolumn{7}{c}{Normalized Write Cycles (NWC)}\\
       &          &  0.0 & 0.1 & 0.3 & 0.5 & 0.7 & 0.9 & 1.0 \\
        \midrule
    \multirow{4}{*}{0.1}
    & \emph{SWIM}      & $\uparrow$    & \textbf{98.49 $\pm$ 0.08} & \textbf{98.56 $\pm$ 0.08} & \textbf{98.57 $\pm$ 0.08} & \textbf{98.57 $\pm$ 0.08} & \textbf{98.57 $\pm$ 0.08} & $\uparrow$\\
    & Magnitude & \multirow{2}{*}{97.96 $\pm$ 0.31} & 98.20 $\pm$ 0.19 & 98.41 $\pm$ 0.12 & 98.50 $\pm$ 0.09 & 98.54 $\pm$ 0.08 & 98.56 $\pm$ 0.08 & 98.58 $\pm$ 0.08\\
    & Random    &               & 98.03 $\pm$ 0.26 & 98.17 $\pm$ 0.21 & 98.30 $\pm$ 0.16 & 98.42 $\pm$ 0.12 & 98.52 $\pm$ 0.09 & $\downarrow$\\
    & In-situ   & $\downarrow$  & 98.39 $\pm$ 0.21 & 98.46 $\pm$ 0.19 & 98.47 $\pm$ 0.17 & 98.48 $\pm$ 0.16 & 98.50 $\pm$ 0.17 & 98.51 $\pm$ 0.17\\
        \midrule
    \multirow{4}{*}{0.15}
    & \emph{SWIM}     & $\uparrow$    & \textbf{98.30 $\pm$ 0.13} & \textbf{98.52 $\pm$ 0.09} & \textbf{98.57 $\pm$ 0.08} & \textbf{98.57 $\pm$ 0.08} & \textbf{98.58 $\pm$ 0.08} & $\uparrow$\\
    & Magnitude & \multirow{2}{*}{96.13 $\pm$ 1.23} & 97.33 $\pm$ 0.56 & 98.14 $\pm$ 0.21 & 98.43 $\pm$ 0.12 & 98.51 $\pm$ 0.10 & 98.56 $\pm$ 0.08 & 98.58 $\pm$ 0.08\\
    & Random    &               & 96.53 $\pm$ 1.04 & 97.20 $\pm$ 0.65 & 97.73 $\pm$ 0.39 & 98.12 $\pm$ 0.23 & 98.45 $\pm$ 0.12 & $\downarrow$\\
    & In-situ   & $\downarrow$  & 96.47 $\pm$ 1.00 & 96.59 $\pm$ 0.82 & 96.69 $\pm$ 0.84 & 96.72 $\pm$ 0.82 & 96.79 $\pm$ 0.85 & 96.84 $\pm$ 0.77 \\
        \midrule
    \multirow{4}{*}{0.2}
    & \emph{SWIM}      & $\uparrow$    & \textbf{98.12 $\pm$ 0.16} & \textbf{98.46 $\pm$ 0.09} & \textbf{98.55 $\pm$ 0.08} & \textbf{98.57 $\pm$ 0.08} & \textbf{98.58 $\pm$ 0.08} & $\uparrow$\\
    & Magnitude & \multirow{2}{*}{94.46 $\pm$ 2.16} & 96.20 $\pm$ 1.11 & 97.65 $\pm$ 0.39 & 98.29 $\pm$ 0.14 & 98.45 $\pm$ 0.10 & 98.54 $\pm$ 0.08 & 98.58 $\pm$ 0.08\\
    & Random    &               & 94.89 $\pm$ 1.90 & 96.13 $\pm$ 1.20 & 97.15 $\pm$ 1.43 & 97.88 $\pm$ 0.71 & 98.38 $\pm$ 0.20 & $\downarrow$\\
    & In-situ   & $\downarrow$  & 95.33 $\pm$ 1.75 & 95.96 $\pm$ 1.36 & 96.42 $\pm$ 1.18 & 96.49 $\pm$ 1.09 & 96.69 $\pm$ 0.94 & 96.82 $\pm$ 0.80\\
        % \midrule
    % \multirow{4}{*}{0.3}
    % & SIWM      & $\uparrow$    & \textbf{96.04 $\pm$ 0.82} & \textbf{97.70 $\pm$ 0.42} & \textbf{98.13 $\pm$ 0.26} & \textbf{98.42 $\pm$ 0.08} & \textbf{98.55 $\pm$ 0.08} & $\uparrow$\\
    % & Magnitude & \multirow{2}{*}{76.30 $\pm$ 7.18} & 85.65 $\pm$ 4.35 & 93.80 $\pm$ 1.80 & 97.00 $\pm$ 0.59 & 97.94 $\pm$ 0.28 & 98.43 $\pm$ 0.12 & 98.58 $\pm$ 0.08\\
    % & Random    &               & 79.31 $\pm$ 3.21 & 85.64 $\pm$ 2.05 & 91.27 $\pm$ 1.07 & 95.42 $\pm$ 0.76 & 97.91 $\pm$ 0.16 & $\downarrow$\\
    % & In-situ   & $\downarrow$  & 77.71 $\pm$ 6.59 & 85.64 $\pm$ 4.24 & 86.24 $\pm$ 4,81 & 88.01 $\pm$ 3.56 & 88.67 $\pm$ 3.20 & 89.13 $\pm$ 3.00\\
    
        \bottomrule
    \end{tabular}
    \vspace{-0.4cm}
    \label{tab:LeNet}
\end{table*}

\section{Experimental Evalutaion}\label{sect:exp}
% \subsection{Justification}
% Justification of second derivative by chi2
In this section, we first define the device variation model we use. Then, we describe a comprehensive study on the MNIST dataset to show the effectiveness of {\em SWIM} over the state-of-the-art under different device variations. After that, we use CIFAR-10 and Tiny ImageNet datasets to show its effectiveness in larger models.

\subsection{Mapping and Impact of Device Variations}\label{sect:model}
This paper is a proof concept to show the effectiveness of {\em SWIM} on temporal variations 
in the programming process, where the variation of each device is independent, so we use a simple yet realistic model to describe it. 

For a weight represented by $M$ bits, let its desired value $\mathcal{W}_{des}$ be:
\begin{equation}
    % \vspace{-0.5cm}
    \mathcal{W}_{des} = \sum_{i=0}^{M-1}{m_i \times 2^i}
    % \vspace{-0.2cm}
\end{equation}
where $m_i$ is the value of the $i^{th}$ bit of the desired weight value. 
We also assume the value programmed on each device is a Gaussian variable of $\mathcal{N}(g, \sigma^2)$ where $g$ is the desired conductance value and $\sigma$ describes the level of uncertainty under device variation. Note that $\sigma$ is independent of $\mathcal{W}_{des}$ according to experimental observations~\cite{feinberg2018making}. 

An $M$-bit weight can be mapped to $M/K$ $K$-bit devices\footnote{Wihtout loss of generality, we assume that M is a multiple of K.}, with the mapped value of the $i^{th}$ ($0\leq i\leq M/K-1$) device $g_i$ as:
\begin{equation}
    % \vspace{-0.5cm}
    g_i = \mathcal{N}\left(\sum_{j=0}^{K -1}{ m_{i\times K + j} \times 2^{j}, \sigma^2 }\right)
    \vspace{-0.2cm}
\end{equation}
Note that negative weights are mapped in a similar manner.

Thus, when a weight is programmed, the actual value $\mathcal{W}_{map}$ mapped on the devices would be:
\begin{align}
\begin{split}
    \mathcal{W}_{map}   & = \sum_{i=0}^{M/K -1}2^{i\times K}{\mathcal{N}\left( g_i, \sigma^2\right) } \\
                        % & = \sum_{i=0}^{M/K -1}2^{i\times K}\mathcal{N}\left(\sum_{j=0}^{K -1}{ m_{i\times K + j} \times 2^{j}, \sigma^2 }\right) \\
                        & = \mathcal{W}_{des} + \sum_{i=0}^{M/K -1}2^{i\times K}{\mathcal{N}(0, \sigma^2) }\\
                        & = \mathcal{W}_{des} +\mathcal{N}\left(0, \sigma^2 \sum_{i=0}^{M/K-1}{2^{i\times K \times 2}}\right) \label{eq:noise}
\end{split}
\vspace{-0.5cm}
\end{align}

In the experiments below, we set $K=4$ as in~\cite{jiang2020device} and follow the above model in simulating the write-verify process. Same as the standard practice discussed in Section~\ref{sec:2.1}, for each weight, we iteratively program the difference between the value on the device and 
the expected value until it is below 0.06. With the inherent randomness, it may take different weights different number of cycles to complete the write-verify: some may not need rewrite at all; while others need a lot. Statistically, the above model results in an average of 
% 5 and a maximum of 
10 cycles over all the weights and a weight variation distribution with $\sigma=0.03$ after write-verify. These numbers are in line 
with those reported in~\cite{shim2020two}, which confirms the validity of our model and parameters.

\subsection{Baselines and Metrics}
In addition to the common practice of writing-verifying all weights the comparison with which is quite trivial, we choose three baselines for {\em SWIM} to compare with: 1) \emph{Random selection}: 
each time we randomly select 
a group of weights from the ones that have not yet been selected to perform write-verify. 2) \emph {Magnitude based approach:} we sort all the weights based on their magnitude and conduct write-verify with the largest ones first. 3) \emph {In-situ training:} retrain the networks on-chip following the same method as that used in~\cite{yao2020fully}. No write-verify is performed. 

Because these methods have different programming mechanisms (write-verify vs. on-device training), we use the total number of write cycles as an indicator for the programming time,
which is fair as writing NVM devices takes far more time than reading them and other operations. For the two write-verify baseline methods and {\em SWIM}, the model in Section~\ref{sect:model} is applied in simulating and counting the number of write cycles. On the other hand, the number of writes in each iteration of in-situ training is equal to the number of weights that are selected for update in that iteration as no write-verify is done. 
To better compare different methods, we normalize the number of 
write cycles with respect to that used to write-verify all the weights in the DNN model under the same setting.  

Note that for {\em SWIM}, random selection and magnitude based selection, $0.0 \leq$ NWC $\leq 1.0 $, but for in-situ training, NWC can exceed $1.0$ because the model can be trained for many iterations and need a large number of writes to update the weights. If we do not do any write-verify or in-situ training, then  
NWC $= 0.0$.  In our experiments, we vary the maximum allowed accuracy drop $\delta A$ for each method and collect the resulting NWC needed. 

All models presented are quantized to the proper data precision and trained to converge on GPU before mapping to nvCiM. This training process is quantization-aware following~\cite{jiang2020device} but does not take device variations into considerations. %Also, based on the experimental data in~\cite{yao2020fully}, we set the standard deviation in the device variation model (Eq.~\ref{eq:noise}) to be $0.1$ ($\sigma = 0.1$) for devices not programmed by write-verify and $\sigma = 0.03$ for devices programmed with write-verify. 
The experiments are conducted on GTX Titan-XP GPUs with the machine learning framework of PyTorch 1.8.1. Considering the randomness in device variations, all results shown in this paper are obtained over 3,000 Monte Carlo runs with verified convergence, and both mean and standard deviation are reported. 

% This model is used in experiments afterwards.

\begin{figure*}
\vspace{-0.6cm}

\begin{minipage}[b]{0.27\linewidth}
    \includegraphics[trim=10 0 30 30, clip, width=\linewidth]{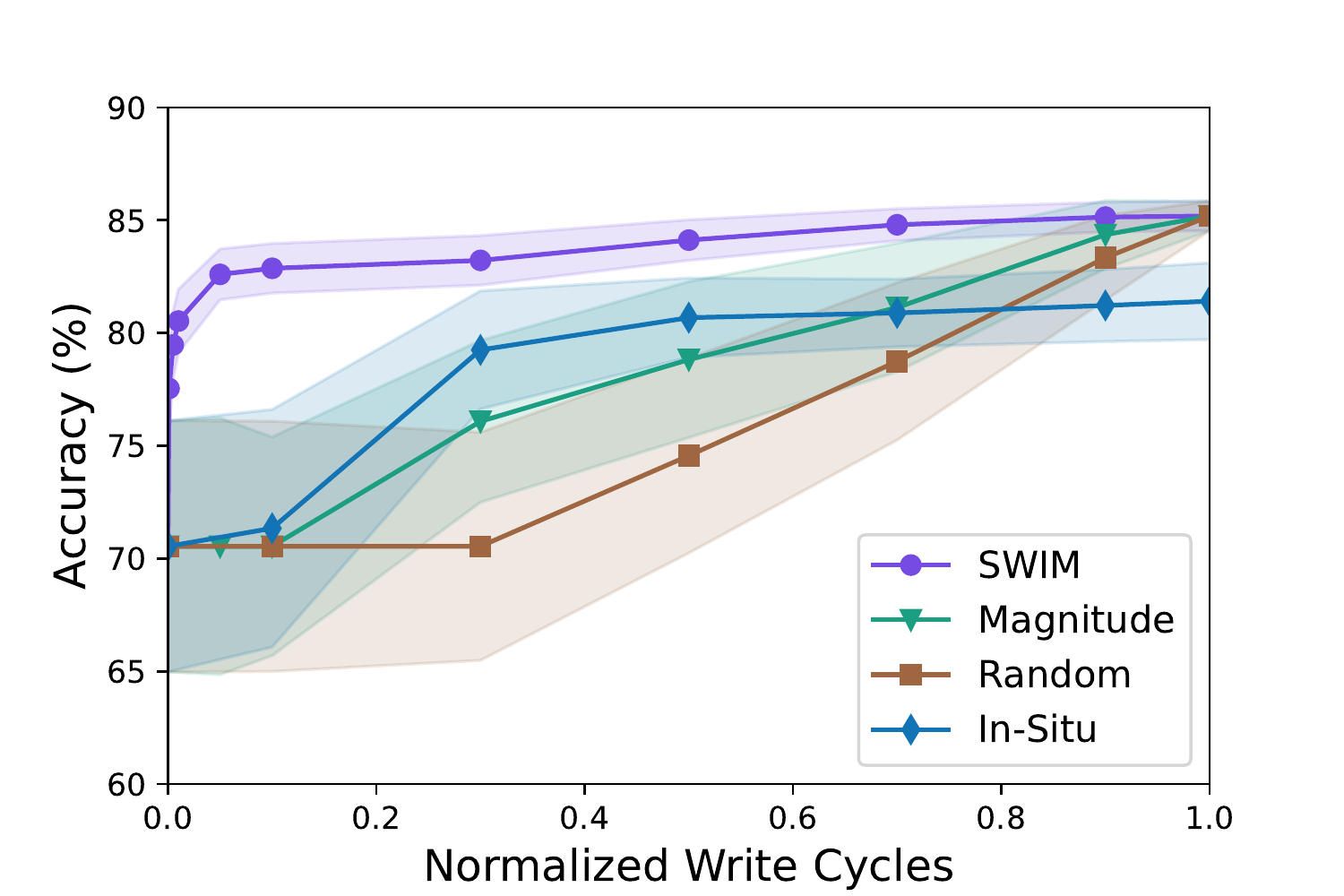}
    \vspace{-0.5cm}
    \subcaption{ConvNet for CIFAR-10}
    \label{fig:ConvNet}
\end{minipage}
\begin{minipage}[b]{0.27\linewidth}
    \includegraphics[trim=8 0 32 30, clip, width=\linewidth]{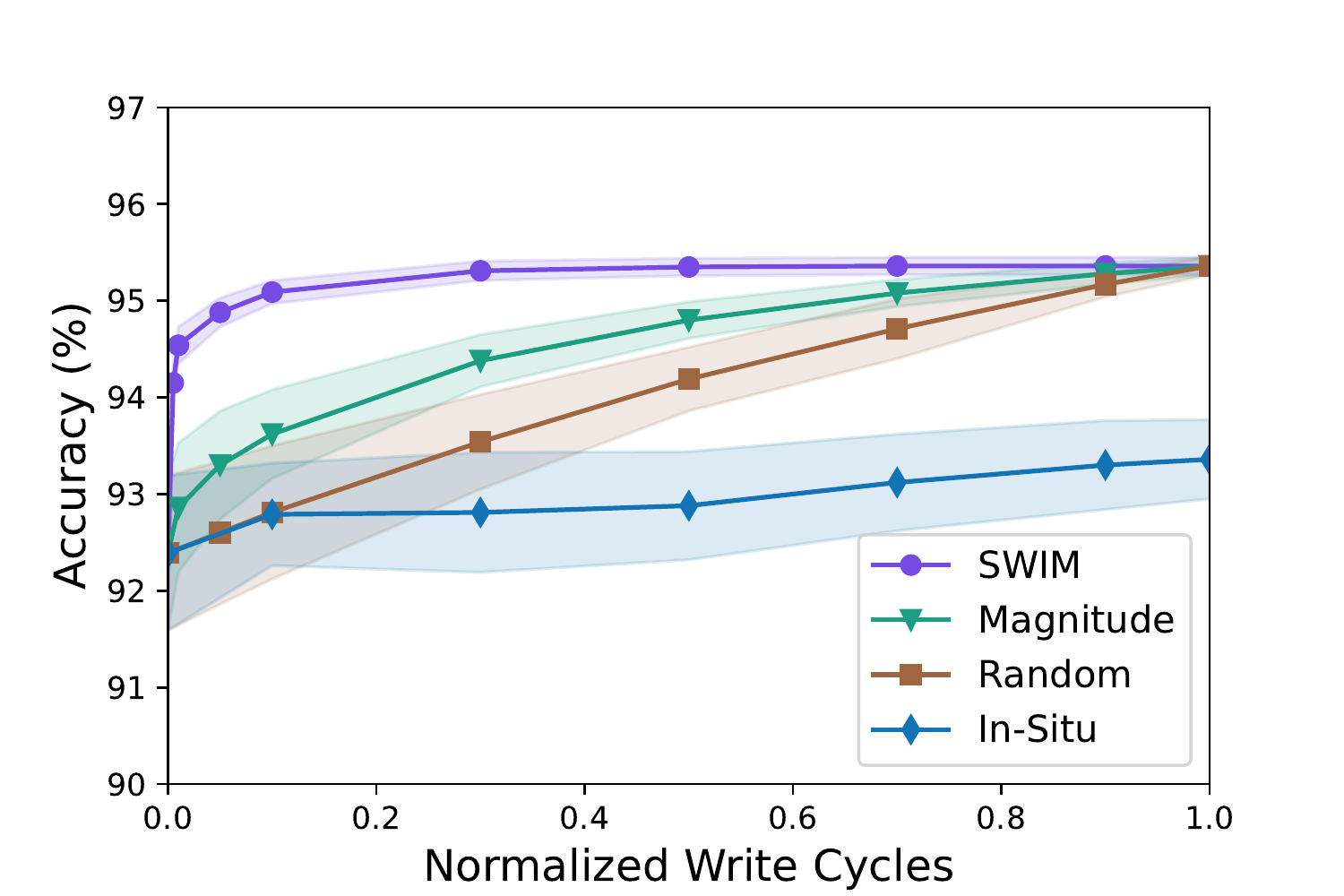}    
    \vspace{-0.5cm}
    \subcaption{ResNet-18 for CIFAR-10}
    \label{fig:Res18}
\end{minipage}
\begin{minipage}[b]{0.27\linewidth}
    \includegraphics[trim=10 0 30 30, clip, width=\linewidth]{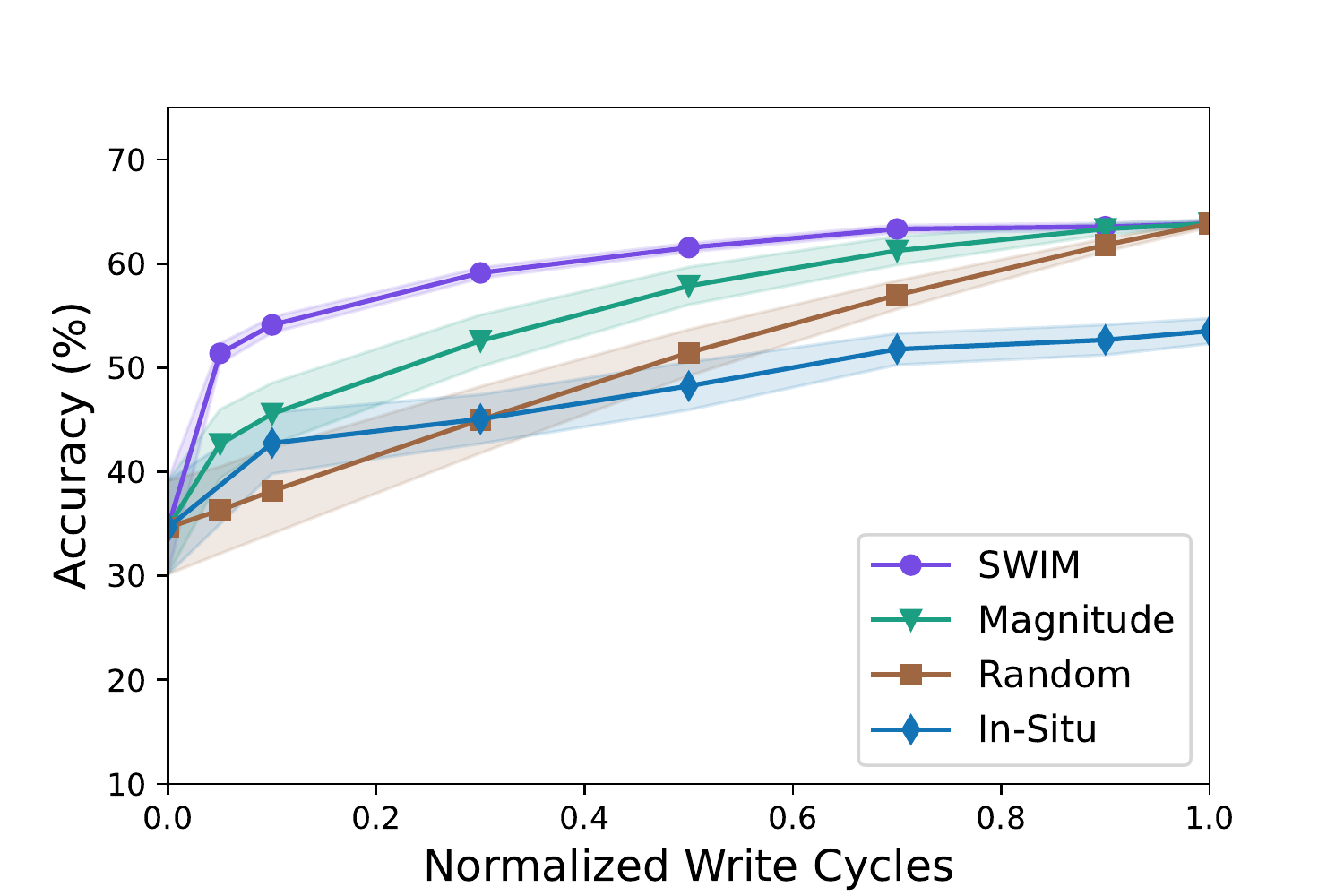}
    \vspace{-0.5cm}
    \subcaption{ResNet-18 for Tiny ImageNet}
    \label{fig:TIN}
\end{minipage}
\vspace{-0.4cm}

\caption{Accuracy v.s. normalized write cycles (NWC) for {\em SWIM} and the baselines on three models: ConvNet and ResNet-18 for CIFAR-10 and ResNet-18 for Tiny ImageNet. Solid lines represent average accuracy and shadowed areas represent standard deviation, over 3,000 Monte Carlo runs using the device variation model. 
% The purple line shows the performance of {\em SWIM}, green the magnitude-based selection, brown random selection and blue in-situ training.
% {\em SWIM} can achieve much lower accuracy drop than three baselines in different normalized write-verify cycles.
}
\vspace{-0.5cm}
% \todo{This figure is generated from testcode/CIM/9874/plot/crossbar.pptx}
\label{fig:all_result}
\end{figure*}

\subsection{Results for MNIST}

We first show the effectiveness of {\em SWIM} on LeNet for the MNIST dataset. Both the weights and activation are quantized to 4 bits. The accuracy of this DNN model without the impact of device variation is 98.68\%. The total number of weights of this model is $1.05\times 10^5$.

Although the typical standard deviation $\sigma$ for device variation model can be assumed to be $0.1$ before write-verify for most devices, certain emerging technologies may lead to higher variations especially before they become mature. To show the broad effectiveness of {\em SWIM}, we compare the performance of {\em SWIM} with the baseline methods over different $\sigma$ values. The results are shown in Table.~\ref{tab:LeNet}. We can see that writing-verifying all the weights can mostly recover the model accuracy (i.e., 98.58\% when NWC $=1.0$ for the three write-verify methods). While all the methods show a decrease in accuracy as NWC decreases, \emph{SWIM} uses significantly fewer NWC than others to attain the same accuracy across all different $\sigma$ values. In addition, it also achieves a significantly lower standard deviation in accuracy than any other method over $3,000$ Monte Carlo runs, indicating that the accuracy would barely fluctuate across different devices. 

Specifically, compared with the conventional practice of writing-verifying all the weights (NWC $=1.0$), with the typical variation ($\sigma=0.1$), {\em SWIM} only needs 50\% of the write cycles (NWC $=0.5$, or $2\times$ speedup) to avoid any accuracy drop. Even with only 10\% of the write cycles (NWC $=0.1$ or $10\times$ speedup), \emph{SWIM} can attain an accuracy drop below $0.1\%$. On the other hand, the magnitude based approach, the random approach, and the in-situ training need an NWC close to 0.5, 0.9, and 0.9, respectively, for that accuracy. This translates to a speedup for {\em SWIM} of $5\times$, $9\times$, and $9\times$, respectively. {\em SWIM} remains effective even when $\sigma$ reaches $0.2$: compared with writing-verifying all the weights, by using 10\% of the write cycles, it can achieve an accuracy drop of less than 0.5\%. To achieve this accuracy, random approach and magnitude based approach need 70\% and 50\% of the write cycles respectively. And even with $10\times$ writes (NWC $= 1.0$), in-situ training cannot achieve the same accuracy, indicating that it still needs more training iterations.  
While not shown in Table.~\ref{tab:LeNet}, in-situ training can fully recover model accuracy (to 98.68\%) using 32 NWC, which means it can achieve higher accuracy than the write-verify methods, but at the cost of a significantly larger number of writes and thus significantly longer programming time, as well as the additional hardware.

\subsection{Results for CIFAR-10}

We now show the effectiveness of {\em SWIM} on the CIFAR-10 dataset with two models ConvNet~\cite{peng2019dnnneurosim} and ResNet-18~\cite{he2016deep}. 
For these two models, both the weights and activation are quantized to 6 bits and $\sigma=0.1$ before write-verify. The accuracy without device variations for ConvNet is 86.07\% and for ResNet-18 is 95.62\%. With device variation and all the weights written-verified, the numbers are 85.19\% and 95.36\%. respectively. The total number of weights for ConvNet and ResNet-18 are $6.40\times 10^6$ and $1.12\times 10^7$, respectively.

Fig.~\ref{fig:ConvNet} shows the comparison between {\em SWIM} and the baselines on ConvNets. Compared with writing-verifying all the weights, 
all the methods except {\em SWIM} see an accuracy drop over 10\% when NWC is 0.1, while {\em SWIM} keeps the accuracy drop below 2.5\%. From this figure, we can clearly see that {\em SWIM} has the smallest standard deviation in accuracy among all the methods, demonstrating its superior robustness. While not shown in Fig.~\ref{fig:ConvNet}, with NWC $= 75$, in-situ training can fully recover model accuracy.

Fig.~\ref{fig:Res18} shows the comparison between {\em SWIM} and the baselines on ResNet-18. Similar conclusions can be drawn here. Compared with writing-verifying all the weights, {\em SWIM} can preserve an accuracy drop of less than 0.5\% using only 10\% of the write cycles, while the other methods result in an accuracy drop of more than 2\% for the same number of write cycles. In-situ training can fully recover model accuracy with 115 NWC. 

\subsection{Experiments on Tiny ImageNet}

Finally, we show the effectiveness of {\em SWIM} on Tiny ImageNet with ResNet-18~\cite{he2016deep}, following the same quantization setting and $\sigma$. The accuracy is 65.50\% without device variation, and 64.84\% with device variation and all weight written-verified. The total number of weights for this model is $1.13\times 10^7$.

Fig.~\ref{fig:TIN} shows the comparison between {\em SWIM} and the baselines on ResNet-18 for Tiny ImageNet. As this is a more challenging task than CIFAR-10, we can see that the accuracy drops for all the methods are larger compared with those in Fig.~\ref{fig:Res18}. Even so, {\em SWIM} can achieve an accuracy less than 3\% lower than that of writing-verifying all the weights using only 10\% of the write cycles, fewest of all the methods. In-situ training can fully recover model accuracy in 155 NWC.

\section{Conclusions}
In this work, contrary to the common practice that write-verify  all the weights of a DNN when mapping it to an nvCiM platform to combat device non-idealities, we show that it is only necessary to write-verify a small portion of them while maintaining the accuracy. As such, the programming time can be drastically reduced. We further introduce {\em SWIM}, which efficiently computes second derivatives that can be used to select weights for write-verify. Experimental results show up to 10x speedup compared with conventional write-verify schemes with little accuracy difference.

\bibliographystyle{ACM-Reference-Format}
\bibliography{M6_References}

\end{document}